%% file: non-anonymous-latex-2027.tex
\documentclass[letterpaper]{article} 
\usepackage[preprint]{aaai2027}  
\usepackage[hyphens]{url}  
\usepackage{graphicx} 
\usepackage{natbib}  
\usepackage{caption} 
\usepackage{amsmath}
\usepackage{amssymb}
\usepackage{booktabs}
\usepackage{multirow}
\usepackage{colortbl}

\title{Generative Embedding Benchmark: \\How Much Information Survives in a Dense Embedding?}
\author{
    Yun Li\textsuperscript{\rm 2},
    Biao Yang\textsuperscript{\rm 1}\corresponding,
    Peixi Wu\textsuperscript{\rm 3},
    Yunhao Zhou\textsuperscript{\rm 1},\\
    Mingzhou Jiang\textsuperscript{\rm 4},
    Wei Yuan\textsuperscript{\rm 1},
    Fan Yang\textsuperscript{\rm 1},
    Wenwu Ou\textsuperscript{\rm 1}
}
\affiliations{
    \textsuperscript{\rm 1}Kuaishou Technology \quad
    \textsuperscript{\rm 2}Fudan University\\
    \textsuperscript{\rm 3}University of Science and Technology of China \quad
    \textsuperscript{\rm 4}Tsinghua University\\
    26213090147@m.fudan.edu.cn, yangbiao@kuaishou.com
}

\begin{document}

\maketitle

\begin{abstract}
Embeddings have emerged as a standard representational interface linking foundation models with downstream systems. Most embedding benchmarks assess representations through discriminative tasks or geometric criteria centered on separability in embedding space. However, strong performance on such evaluations does not establish whether content compressed into an embedding remains accessible to a downstream generator. To address this gap, we introduce the \textbf{Generative Embedding Benchmark (GEB)}, in which a decoder answers questions using only a frozen embedding and question text, without access to the original image or intermediate visual features. Answer quality under this readout measures \emph{generative information}: the answer-relevant content recoverable from an embedding. GEB includes a curated visual-question-answering dataset with a 1{,}800-item development split and a held-out 900-item test split covering natural images, scene text, and visual documents. Using a common decoder and training recipe, we evaluate seven public embedding models in visual-only and vision-language joint modes. On the test set, visual-only scores range from 28.25 to 33.21; with image--question joint encoding, all five VLM-based embedding models score higher, and the best reaches 65.56. Matched embeddings also outperform text-only inputs, zero embeddings, and shuffled embeddings. Natural-image information is much easier to recover than scene text or visual-document information, while a Qwen3-VL-2B reference with access to the original image reaches 84.30. Together, these results show that generative readout exposes information bottlenecks that separability-based evaluation does not capture. Homepage at \url{https://github.com/LimitedMouse/Generative-Embedding-Benchmark}.
\end{abstract}

\section{Introduction}

Multimodal embeddings have become a standard representation interface between foundation models and downstream systems. Images, text, and image--text pairs are compressed into dense vectors that can be stored, compared, retrieved, and reused by downstream components. Beyond serving as retrieval keys, embeddings are increasingly explored as compact interfaces for conditioning generators, representing latent memory, and passing information across model boundaries~\cite{sastre2025memory,ge2023icae,xrag2024}. For these uses, the central question is not only whether an embedding can identify a matching item, but how much semantic information remains accessible from the vector to a downstream generator.

\begin{figure}[t!]
\centering
\includegraphics[width=\columnwidth]{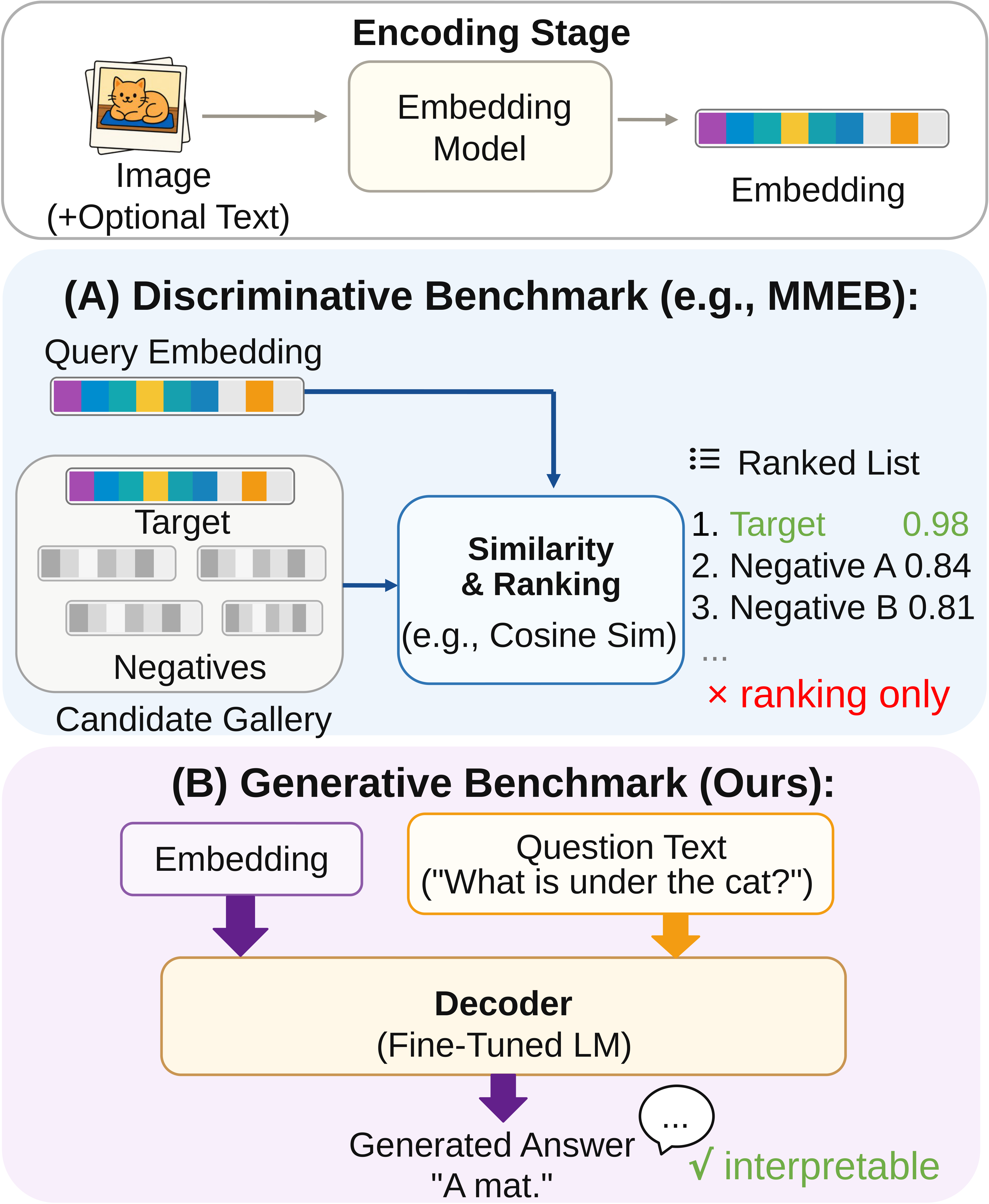}
\caption{Discriminative benchmarks measure separability against a candidate gallery; GEB measures answer-relevant content accessible through generative readout.}
\label{fig:protocol}
\end{figure}

Most existing embedding benchmarks evaluate representations through discriminative tasks or geometric criteria built around separability in embedding space~\cite{conneau2018senteval,muennighoff2023mteb,xiao2025mieb}. Multimodal suites such as MMEB~\cite{jiang2024vlm2vec} and MMEB-V2~\cite{meng2025vlm2vecv2} provide representative gallery-based formulations. These benchmarks measure task-relative discriminative utility effectively, but leave open how much content remains available in an embedding for generation.

We use \emph{generative information} to denote the answer-relevant content that a fixed generative readout can recover from an embedding. An embedding may perform well on separability-based evaluations while still discarding text, layout, counts, attributes, relations, or other details that the evaluated distinction does not require. Success on a predefined discriminative task therefore does not establish that the evidence needed for generation remains recoverable.

Gallery-based evaluation makes this gap especially clear. Its scores are candidate-relative, varying with gallery size and distractor composition. More fundamentally, ranking rewards the information needed to place the annotated target ahead of its alternatives; retained content beyond that distinction receives no additional credit. Successful separation therefore does not necessarily imply that the underlying content is recoverable. Figure~\ref{fig:protocol} summarizes the contrast between gallery-based separability and generative readout.

We propose the \textbf{Generative Embedding Benchmark (GEB)} to evaluate embeddings through generative readout. Given an embedding and the textual question, a decoder generates the answer without ranking over an embedding gallery. For each frozen embedding model and encoding mode, GEB trains a separate decoder using the same readout design, training data, optimization budget, supervised answer tokens, and evaluation procedure. A lightweight input adapter handles differences in embedding dimensionality.

The two encoding modes distinguish reusable visual information encoded before a question is known from query-conditioned information selected for a known question. In \emph{visual-only} mode, the embedding model encodes only the image, while the decoder receives the question separately as text. This tests whether the image embedding preserves evidence useful for questions not provided at encoding time. In \emph{vision-language joint} mode, the embedding model encodes the image together with the question, while the decoder also receives the same question as text. This tests whether the resulting embedding carries the query-relevant information needed to generate the answer. In both modes, the image itself is withheld from the decoder, making the embedding its only visual channel.

GEB combines the generative readout protocol with a curated visual-question-answering dataset comprising development and held-out test splits. The development split is used for protocol selection, decoder sensitivity, and ablations, while the test split is reserved for final model comparison. Together, the two splits cover natural images, scene text, and visual documents. We evaluate seven public embedding models, spanning VLM-based embedders and contrastive image--text baselines such as CLIP~\cite{radford2021clip} and SigLIP~\cite{zhai2023siglip}. Matched embeddings substantially outperform non-informative and shuffled controls, and the two encoding modes produce markedly different model scores and orderings. A VLM reference receiving the original image remains substantially stronger, especially on scene text and visual documents.

In summary, this paper makes three contributions:
\begin{itemize}
\item We formulate generative information as the answer-relevant content exposed by generative readout, providing an evaluation axis beyond task-relative separability.
\item We introduce GEB, comprising visual-only and vision-language joint encoding modes together with development and test sets spanning natural images, scene text, and visual documents.
\item Across seven embedding models, controlled interventions show that GEB scores depend on matched, sample-specific embeddings, while category and MMEB-V2 comparisons reveal evaluation differences not captured by ranking alone.
\end{itemize}

\section{Related Work}

\subsection{Multimodal Embedding Models}

Multimodal embedding models have evolved from contrastive image--text encoders to general-purpose embedders built on vision-language models. CLIP~\cite{radford2021clip} and SigLIP~\cite{zhai2023siglip} learn aligned image and text representations through large-scale contrastive training. More recent systems adapt multimodal language models for embedding. VLM2Vec~\cite{jiang2024vlm2vec} converts instruction-following vision-language models into multimodal embedders through contrastive training, while VLM2Vec-V2~\cite{meng2025vlm2vecv2} extends the framework to videos and visual documents. Qwen3-VL-Embedding and Qwen3-VL-Reranker~\cite{li2026qwen3vlembedding} form a unified framework for multimodal retrieval and reranking, while UME-R1~\cite{lan2025umer1} and Embed-RL~\cite{jiang2026embedrl} introduce reasoning-oriented training for multimodal representations. Together, these systems span contrastive dual encoders, instruction-tuned VLM embedders, and reasoning-oriented embedding models. Our experiments include representatives from these model families under a common generative evaluation protocol.

\subsection{Discriminative Embedding Evaluation}

Embedding benchmarks commonly evaluate downstream utility through discriminative tasks and geometric criteria defined over embedding space. SentEval~\cite{conneau2018senteval}, MTEB~\cite{muennighoff2023mteb}, MMTEB~\cite{enevoldsen2025mmteb}, and MIEB~\cite{xiao2025mieb} collectively span downstream tasks such as semantic similarity, classification, clustering, retrieval, and reranking across text, multilingual, and image settings. In multimodal evaluation, the MMEB series~\cite{jiang2024vlm2vec,meng2025vlm2vecv2,huang2026mmebv3} reformulates heterogeneous tasks as ranking the correct target in a candidate gallery. The original MMEB covers classification, VQA, multimodal retrieval, and visual grounding, while later versions extend the formulation to video and visual-document tasks. Across these settings, separability in embedding space characterizes task-relative discriminative utility: the embedding must support the distinctions required by a downstream objective. GEB studies a complementary property by using generation to measure the answer-relevant content accessible from an embedding.

\subsection{Probing and Generative Readout}

Representation probing asks what information can be read from a frozen representation. Classical probes train supervised classifiers to predict predefined linguistic or structural properties~\cite{conneau2018probing,hewitt2019structural}. Work on control tasks and usable information further relates probe results to the capacity of the readout~\cite{hewitt2019control,xu2020usable}. From this perspective, probing provides an operational measurement of what a representation makes accessible under a specified predictor family. GEB follows the same principle with a generative readout: each question specifies the information of interest, and an embedding-conditioned decoder generates the answer.

Beyond probing, another line of work asks whether compact representations can directly support generation. GEIA~\cite{li2023geia} and Vec2Text~\cite{morris2023vec2text} recover text from sentence embeddings. Memory Tokens~\cite{sastre2025memory}, SONAR~\cite{duquenne2023sonar}, and bidirectional reconstruction objectives~\cite{su2025bidirectional} learn representations that support reconstruction or translation, while AutoCompressor~\cite{chevalier2023autocompressor}, ICAE~\cite{ge2023icae}, and xRAG~\cite{xrag2024} use compact continuous states to condition subsequent generation. In vision, ClipCap~\cite{mokady2021clipcap}, CapDec~\cite{nukrai2022capdec}, and DeCap~\cite{li2023decap} connect frozen CLIP representations to language models for caption generation, while MM-GEM~\cite{ma2024mmgem} jointly optimizes embedding and generation objectives in a single multimodal model. Visual information loss has also been studied within VLMs through reconstruction and neighborhood changes across connector projections~\cite{lost2025embeddings}, as well as through reconstruction from CLIP embeddings~\cite{dorazio2025clipinv}. Together, these works show that compact representations can support generation and reconstruction. GEB uses this capability as an evaluation mechanism, holding independently trained embedding models fixed and comparing them under a shared readout.

\section{Method}
\label{sec:method}

\subsection{Protocol Overview}

GEB evaluates a frozen embedding model through generative readout. This design tests not merely whether embeddings support discrimination, but whether the content they retain can be recovered for conditional generation. For each model $m$, we train a separate embedding-conditioned decoder $D_m$ that answers visual questions without access to the original image tokens or any intermediate visual features. The embedding is therefore the only channel through which image information reaches the decoder. By keeping the decoder architecture and training protocol fixed across models, answer quality serves as an operational measure of the answer-relevant information accessible to the generative readout.

\begin{figure}[t!]
\centering
\includegraphics[width=\columnwidth]{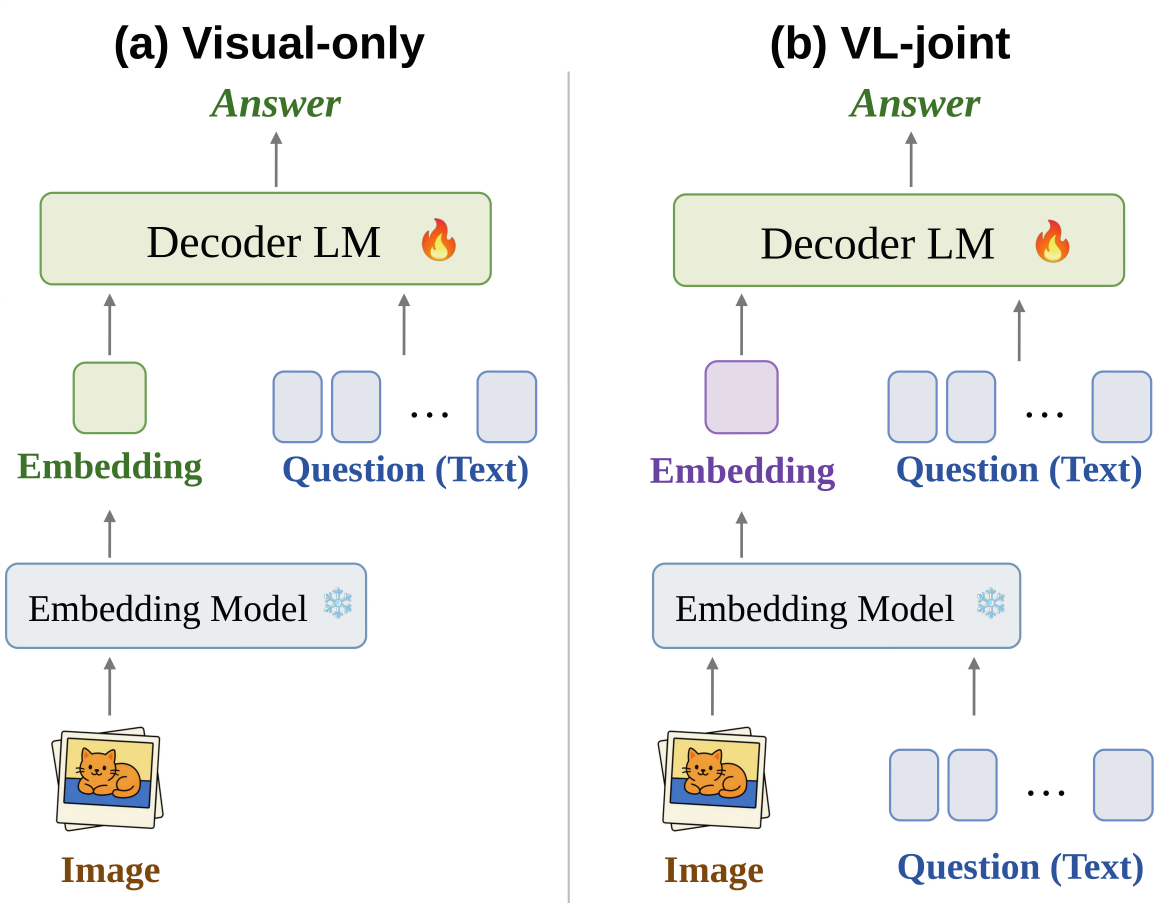}
\caption{GEB encoding modes. The decoder always receives the textual question but not the original image; VL-joint also conditions the embedding on the question.}
\label{fig:encoding_modes}
\end{figure}

Figure~\ref{fig:encoding_modes} illustrates the two GEB encoding modes. In both modes, the decoder receives the textual question; visual-only produces a question-agnostic image embedding, whereas vision-language joint (VL-joint) produces an embedding conditioned jointly on the image and question.

In \emph{visual-only} mode, the frozen embedding model encodes each image $x_i$ once, independently of any downstream question:
\begin{equation}
\mathbf{e}_i = E_m(x_i), \qquad
\hat{a}_{i,q} = D_m\big(A_m(\mathbf{e}_i), q\big),
\end{equation}
where $A_m$ is a lightweight input adapter that maps model-specific embeddings to the decoder input space, and the decoder receives the question $q$ as ordinary text. The same image embedding can be reused across questions, so successful generation requires it to preserve visual information useful for questions unavailable at encoding time.

In \emph{vision-language joint} mode, the embedding model jointly encodes the image together with the current question,
\begin{equation}
\mathbf{e}_{i,q} = E_m(x_i,q), \qquad
\hat{a}_{i,q} = D_m\big(A_m(\mathbf{e}_{i,q}), q\big).
\end{equation}
The decoder also receives the same question as text. In this mode, the embedding is evaluated as the source of visual evidence, rather than as a complete encoding of the generation instruction. Thus, it evaluates whether a question-conditioned embedding makes the query-relevant information needed to generate the answer accessible to the decoder. Models with native joint encoding produce a single embedding vector. CLIP and SigLIP are included only as contrastive references; because they lack native joint encoding, their VL-joint variant uses a two-slot late-fusion interface that maps separate image and text embeddings into the decoder.

\subsection{Decoder Architecture}

The decoder consists of two trainable components: a lightweight adapter that maps the embedding into the LM hidden space, and a decoder-only language model (LM) that generates the answer.

The adapter is a small MLP. A frozen embedding $\mathbf{e}\in\mathbb{R}^{d}$ is layer-normalized and passed through two linear layers with one GELU in between,
\begin{equation}
\mathbf{h} = W_2\,\text{GELU}\big(W_1\,\text{LN}(\mathbf{e})\big),
\end{equation}
where $W_1 \in \mathbb{R}^{d_h \times d}$ projects the embedding to a fixed width $d_h{=}1024$ and $W_2 \in \mathbb{R}^{d_{\text{dec}} \times d_h}$ maps it to the LM hidden size. The projected vector $\mathbf{h}$ replaces the input embedding of a special \texttt{<EMBED\_PAD>} token at the start of the first user turn. The question and conversation context remain ordinary text inputs.

We use Qwen3-0.6B~\cite{yang2025qwen3} as the decoder-only LM and fine-tune it under its native chat template. The language-modeling loss is computed only on assistant tokens and masks system, user, and template tokens from the loss. No original image or intermediate visual features are exposed to the decoder.

\subsection{Training Protocol}

For each embedding model and encoding mode, we train a separate decoder using the same recipe and supervised-token budget:
\begin{itemize}
\item \textbf{Data}: LLaVA-NeXT 738K multi-turn conversations~\cite{liu2024llavanext}, rendered in the native chat template with the full multi-turn context per image.
\item \textbf{Optimization}: 1 epoch; effective batch size 512 (8 GPUs $\times$ 16 gradient accumulation $\times$ 4 per device); learning rate $1 \times 10^{-4}$ with cosine decay.
\item \textbf{Sequence length}: dynamic padding within each batch, with a truncation limit of 32{,}768 tokens.
\item \textbf{Trainable parameters}: the adapter and all LM weights. The embedder $E$ is never updated.
\end{itemize}

In visual-only mode, one image embedding is shared across the full conversation and all assistant turns are supervised. In VL-joint mode, we expand each conversation into turn-level training instances. The embedding is computed from the image and current user question for each assistant turn; prior turns remain available to the decoder as textual context, and loss is applied only to the current answer.

The recipe is fixed before comparison and is not tuned for any individual embedding model. The adapter accommodates differences in embedding dimensionality, while the decoder LM and training recipe remain unchanged. Although a separate decoder is trained for each embedding model, all models share the same decoder design, training data, optimization budget, supervised answer tokens, and evaluation procedure. Under this standardized readout protocol, answer quality provides an operational measure of the answer-relevant information recoverable from each embedding.

\section{Benchmark Construction}

\begin{table}[t!]
\centering
\small
\begin{tabular*}{\columnwidth}{@{\extracolsep{\fill}}llrr@{}}
\toprule
Category & Source & Test & Dev \\
\midrule
\multirow{3}{*}{Natural image} & MME & 160 & 320 \\
 & CV-Bench & 80 & 160 \\
 & RealWorldQA & 60 & 120 \\
\midrule
\multirow{3}{*}{Scene text} & TextVQA & 180 & 360 \\
 & OCRBench & 80 & 160 \\
 & MME & 40 & 80 \\
\midrule
\multirow{4}{*}{Visual document} & ChartQA & 100 & 200 \\
 & DocVQA & 120 & 240 \\
 & InfoVQA & 70 & 140 \\
 & OCRBench & 10 & 20 \\
\midrule
Total & & 900 & 1{,}800 \\
\bottomrule
\end{tabular*}
\caption{GEB dataset composition. The test split contains 300 items per category; the development split uses twice each source quota.}
\label{tab:dataset_composition}
\end{table}

The GEB dataset contains a 1{,}800-item development split and a 900-item test split. Each split is balanced across three semantic categories. \textbf{Natural image} items ask about objects, counts, attributes, spatial relations, landmarks, and scenes. \textbf{Scene text} items require reading text in photographs, signs, posters, and handwriting. \textbf{Visual document} items cover charts, forms, receipts, and infographics. Together, the categories span general visual recognition, text reading, and structured document understanding.

\subsection{Candidate Pool and Answerability}

Candidate items are drawn from the public tasks shown in Table~\ref{tab:dataset_composition}, as implemented in LMMs-Eval~\cite{fu2023mme,tong2024cambrian,xai2024realworldqa,singh2019textvqa,liu2024ocrbench,mathew2021docvqa,masry2022chartqa,mathew2022infographicvqa}. TextVQA, DocVQA, and InfographicVQA use their validation splits; all other sources use their test splits. We retain candidates that fall within the predefined semantic scope and that Gemini-3-Flash~\cite{google2025gemini3flash} answers correctly under the original task metric. The GEB dataset is therefore a stratified sample of a reference-VLM-answerable pool rather than of the full source benchmarks. Depending on the source, this answerability filter retains 39.4--95.6\% of in-scope items; complete per-source attrition counts are reported in Table~\ref{tab:attrition} and the released metadata.

The semantic rules remove content outside the intended categories or duplicated across source tasks. In particular, CV-Bench retains 2D counting, 2D relation, and 3D depth questions, while excluding 3D distance. OCRBench excludes digit-string recognition, non-semantic text recognition, handwritten mathematical expressions, and subsets already represented by TextVQA, ChartQA, DocVQA, or InfographicVQA; its visual-document contribution is key-information extraction. Candidate selection does not use GEB decoder outputs, evaluated embedding models, pilot scores, control conditions, or model disagreement.

\subsection{Development and Test Splits}

We construct the development and test splits by stratified random sampling from the answerable candidate pool. The test split contains 300 items per category, and the development split contains 600, with exactly twice each test-set source quota. The unit of assignment is an image-identity group rather than an individual question: all questions associated with the same image are kept in one set. Identity is determined from decoded-image hashes, augmented with source-specific perceptual grouping for CV-Bench and RealWorldQA to keep duplicated frames and image variants together. A leakage audit confirms that no image-identity group crosses the two splits: row-key overlap, image-group overlap, exact decoded-image overlap, and identical cross-split perceptual hashes are all zero (Table~\ref{tab:attrition}).

The development split is used for decoder and protocol development. The test split is reserved for the final model comparisons and is not used to revise the protocol. The released manifests record source membership, category labels, image-group identifiers, and checksums so that the two sets can be reproduced independently.

\begin{table}[t]
\centering
\small
\setlength{\tabcolsep}{4pt}
\begin{tabular*}{\columnwidth}{@{\extracolsep{\fill}}lrrrr@{}}
\toprule
Source & Raw & In-scope & Answerable & Selected \\
\midrule
CV-Bench      & 2{,}638 & 2{,}038 & 1{,}795 & 240 \\
MME           & 2{,}374 & 2{,}284 & 2{,}146 & 600 \\
RealWorldQA   &   765 &   765 &   644 & 180 \\
TextVQA       & 5{,}000 & 2{,}900 & 2{,}386 & 540 \\
OCRBench      & 1{,}000 &   550 &   511 & 270 \\
ChartQA       & 2{,}500 & 2{,}500 &   984 & 300 \\
DocVQA        & 5{,}349 & 5{,}349 & 5{,}115 & 360 \\
InfographicVQA& 2{,}801 & 2{,}801 & 2{,}474 & 210 \\
\bottomrule
\end{tabular*}
\caption{Per-source candidate attrition. \emph{Raw} is the source split size; \emph{In-scope} applies the semantic rules; \emph{Answerable} retains items answered correctly by Gemini-3-Flash; and \emph{Selected} is the total placed in the development and test splits.}
\label{tab:attrition}
\end{table}

\subsection{Metrics}
\label{sec:metrics}

We conduct evaluation using the LMMs-Eval framework~\cite{zhang2025lmmseval}. Our GEB task wrappers retain each source task's original answer-processing procedures and generation settings, including output-length budgets, and map each response to a per-item score in $[0,1]$. MME, CV-Bench, RealWorldQA, and OCRBench use binary correctness after task-specific answer parsing; for OCRBench, a prediction is considered correct if the normalized reference answer appears in the normalized prediction. TextVQA uses its standard VQA consensus score~\cite{singh2019textvqa}, ChartQA uses relaxed accuracy~\cite{masry2022chartqa}, and DocVQA and InfographicVQA use ANLS (Average Normalized Levenshtein Similarity).

We compute all results by directly averaging per-item scores. On the test set, the overall score is averaged across all 900 items, while each category score is averaged across its 300 items. Thus, source tasks are weighted by their number of selected items rather than equally at the task level. All results are reported as percentages.

\section{Experiments}
\label{sec:experiments}

The evaluation proceeds in five stages. First, we define the model set, encoding modes, and evaluation protocol. Next, the readout design is fixed on the development set before the models are compared on the held-out test set. Finally, matched-information controls and comparison with MMEB-V2 assess whether the scores rely on aligned embeddings and whether the resulting ranking agrees with an established benchmark.

\subsection{Embedding Models and Evaluation Protocol}
\label{sec:embedding_models}

\textbf{Models.}
We evaluate seven public embedding models under the shared decoder protocol. Five are instruction-tuned multimodal embedding models built on vision-language models, while CLIP ViT-L/14 and SigLIP SO400M serve as contrastive image--text references~\cite{li2026qwen3vlembedding,meng2025vlm2vecv2,lan2025umer1,jiang2026embedrl,radford2021clip,zhai2023siglip}. Qwen3-VL-Embedding and VLM2Vec pool the last valid token of the final hidden state. UME-R1 and Embed-RL use the hidden state at a dedicated embedding token, while CLIP and SigLIP use their projected pooled representations. Every vector is L2-normalized before entering the adapter.

\textbf{Encoding modes.}
We use the visual-only and VL-joint encoding modes defined in Section~\ref{sec:method}. Visual-only computes one embedding per image and reuses it across questions. VL-joint computes a new embedding from the image and current question for each turn. Both modes use identical answer targets and supervised-token budgets.

\textbf{Evaluation protocol.}
We select all protocol configurations on the development set and reserve the held-out test set for final model comparison and control experiments. All test scores are recomputed from the 900 per-item records using the metrics in Section~\ref{sec:metrics}. Bootstrap uncertainty analyses, including paired control gaps, are reported in the supplementary material.

\subsection{Protocol and Decoder Ablations}
\label{sec:protocol_ablations}

A preliminary development study selects current-question joint encoding while retaining the available textual history at the decoder; the supplementary material reports the full comparison. We then evaluate decoder capacity and second-stage tuning on the fixed 1{,}800-item development set, holding the Qwen3-VL-Embedding-2B visual-only representation fixed (Table~\ref{tab:dev_decoder}).

\begin{center}
\begin{minipage}{\columnwidth}
\centering
\small
\setlength{\tabcolsep}{4pt}
\begin{tabular*}{\columnwidth}{@{\extracolsep{\fill}}llr@{}}
\toprule
Decoder & Recipe & Dev \\
\midrule
Qwen3-0.6B & 1-stage SFT & 34.80 \\
Qwen3-0.6B & 2-stage SFT & 34.20 \\
Qwen3-1.7B & 1-stage SFT & 36.76 \\
Qwen3-1.7B & 2-stage SFT & 37.02 \\
\bottomrule
\end{tabular*}
\captionof{table}{Decoder sensitivity on the development set using Qwen3-VL-Embedding-2B in visual-only mode. The 1-stage recipe is trained on LLaVA-NeXT 738K; the 2-stage recipe adds SFT on 60K format-aligned examples with learning rate $10^{-5}$. Scores are overall percentages.}
\label{tab:dev_decoder}
\end{minipage}
\end{center}

\begin{table*}[t!]
\centering
\setlength{\tabcolsep}{9pt}
\begin{tabular}{@{}lr@{\hspace{18pt}}rrrr@{\hspace{18pt}}rrrr@{}}
\toprule
& & \multicolumn{4}{c}{Visual-only} & \multicolumn{4}{c}{VL-joint} \\
\cmidrule(lr){3-6}\cmidrule(lr){7-10}
Model & Dim & Overall & Natural & Scene & Doc & Overall & Natural & Scene & Doc \\
\midrule
\rowcolor{gray!12}\multicolumn{10}{@{}l}{\textit{Contrastive Embedding Models}} \\
CLIP ViT-L/14 & 768 & 28.25 & 58.00 & 13.03 & 13.73 & 4.97 & 10.00 & 3.33 & 1.57 \\
SigLIP SO400M & 1152 & 30.58 & 60.33 & 18.03 & 13.36 & 21.99 & 47.33 & 12.87 & 5.77 \\
\midrule
\rowcolor{gray!12}\multicolumn{10}{@{}l}{\textit{VLM-based Embedding Models}} \\
Embed-RL-2B & 2048 & 29.21 & 55.67 & 20.63 & 11.32 & 45.91 & 62.33 & 38.67 & 36.73 \\
VLM2Vec-V2 & 1536 & 29.26 & 61.00 & 11.60 & \textbf{15.19} & 46.09 & 58.33 & 39.67 & 40.28 \\
UME-R1-2B & 1536 & 31.21 & \textbf{62.67} & 16.63 & 14.34 & 51.45 & 62.67 & 49.13 & 42.56 \\
Qwen3-VL-Embedding-2B & 2048 & 32.92 & 61.67 & \textbf{23.67} & 13.42 & 44.87 & 50.67 & 41.33 & 42.62 \\
Qwen3-VL-Embedding-8B & 4096 & \textbf{33.21} & \textbf{62.67} & 23.07 & 13.88 & \textbf{65.56} & \textbf{81.67} & \textbf{59.37} & \textbf{55.64} \\
\midrule
\rowcolor{gray!12}\multicolumn{10}{@{}l}{\textit{VLM Reference}} \\
\textit{Qwen3-VL-2B} & \textit{--} & \textit{--} & \textit{--} & \textit{--} & \textit{--} & \textit{84.30} & \textit{81.67} & \textit{85.60} & \textit{85.63} \\
\bottomrule
\end{tabular}
\caption{GEB scores on the test set (\%). Bold marks the best embedding-model score within each encoding mode. CLIP and SigLIP use separate image and text vectors in VL-joint.}
\label{tab:main_results}
\end{table*}

Across the four configurations, the 1.7B decoder improves over the selected 0.6B one-stage baseline by 1.95--2.22 points. Because this gain is modest and our goal is a controlled, low-capacity readout rather than the highest development score, we use the Qwen3-0.6B one-stage decoder for every main comparison. Preliminary studies of joint-input construction, LoRA versus full fine-tuning, and training-data scale are reported in the supplementary material. Once these choices are fixed, the test set is used only for the model comparison in Section~\ref{sec:test_results} and the controls in Section~\ref{sec:control_experiments}.

\subsection{Main Results}
\label{sec:test_results}

With the protocol frozen, we evaluate all seven embedding models on the test set (Table~\ref{tab:main_results}). Qwen3-VL-2B~\cite{bai2025qwen3vl}, which receives the original image tokens instead of an embedding, is included only as a calibration reference.

\textbf{Reusable visual retention.}
Visual-only scores occupy a narrow range from 28.25 to 33.21. Qwen3-VL-Embedding-8B and 2B obtain 33.21 and 32.92, respectively, making them nearly tied under visual-only readout. Natural-image content is substantially more accessible than scene text or visual-document content: the best visual-only scores are 62.67, 23.67, and 15.19 across the three categories.

\textbf{Query-conditioned readout.}
Every VLM-based embedding model improves under VL-joint. Qwen3-VL-Embedding-8B rises from 33.21 to 65.56 and leads every category; UME-R1-2B and VLM2Vec-V2 follow at 51.45 and 46.09. Because VL-joint also uses turn-level training instances, this gain reflects the complete mode-specific pipeline rather than encoder conditioning alone. The image-token Qwen3-VL-2B reaches 84.30 as a calibration reference.

\begin{figure*}[t!]
\centering
\includegraphics[width=0.94\textwidth]{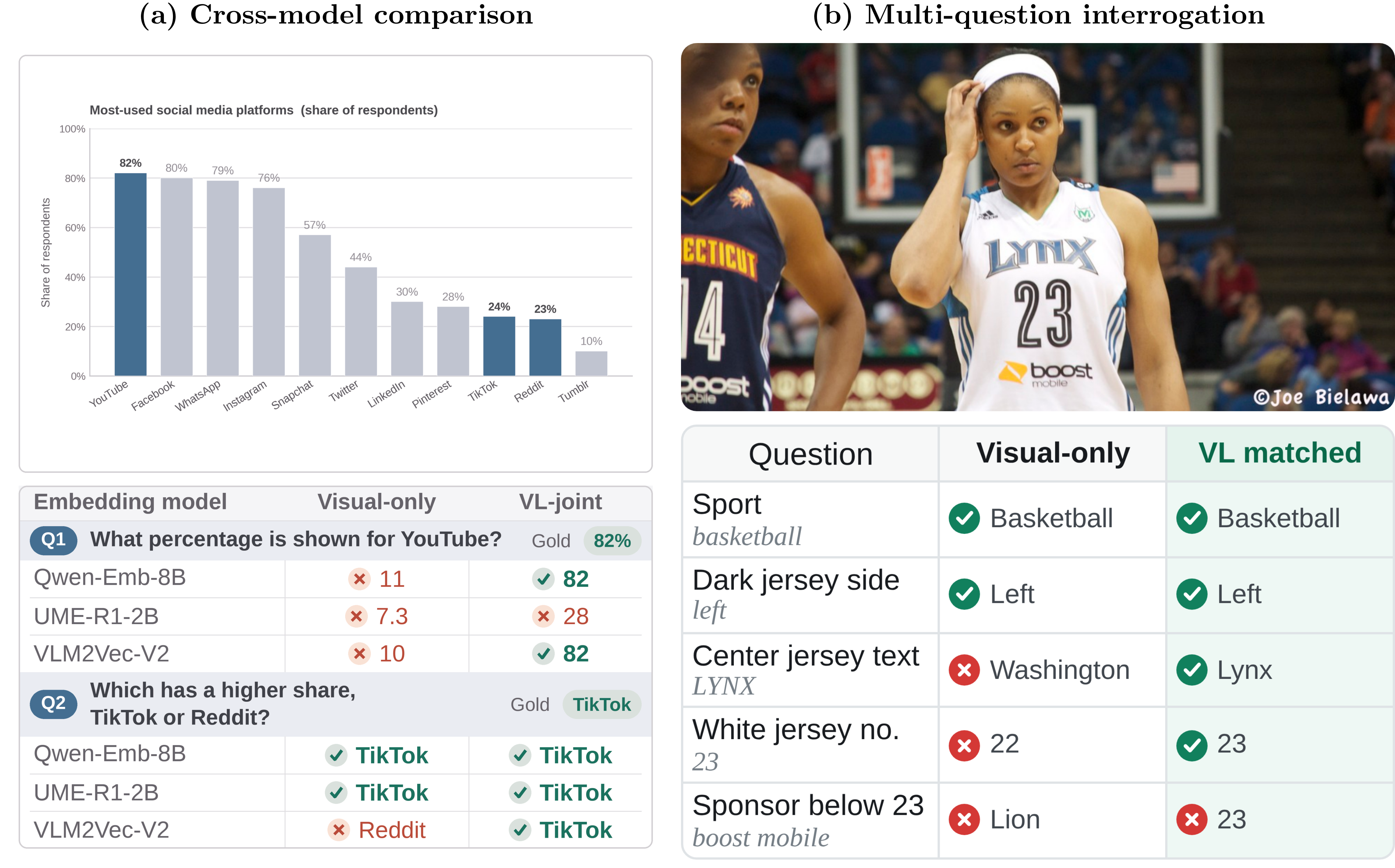}
\caption{Two qualitative views of generative readout: (a) cross-model answers to two questions; (b) multi-question interrogation of Qwen3-VL-Embedding-8B. Italic text in (b) denotes reference answers.}
\label{fig:qualitative_views}
\end{figure*}

\textbf{Cross-model qualitative comparison.}
Figure~\ref{fig:qualitative_views}(a) shows that VL-joint recovers the exact YouTube value 82 for Qwen3-VL-Embedding-8B and VLM2Vec-V2, while UME-R1 reverses the digits to 28; all three visual-only readouts miss it. VLM2Vec-V2 also corrects ``Reddit'' to ``TikTok'' on the comparison question, revealing differences in exact-value and comparative readout.

\textbf{Multi-question information interrogation.}
Figure~\ref{fig:qualitative_views}(b) interrogates one image with five questions. The reused visual-only embedding recovers scene and spatial information but misses jersey text, number, and sponsor details; VL-joint recovers the large text and number but not the smaller text.

\subsection{Control Experiments}
\label{sec:control_experiments}

Table~\ref{tab:controls} tests whether GEB depends on sample-specific embedding information beyond decoder language priors. Visual-only compares matched, text-only, zero, and shuffled inputs; in VL-joint, the decoder retains the correct question while only the encoder input is perturbed.

\begin{center}
\begin{minipage}{\columnwidth}
\centering
\small
\setlength{\tabcolsep}{4pt}
\begin{tabular*}{\columnwidth}{@{\extracolsep{\fill}}llrr@{}}
\toprule
Mode & Condition & Overall & $\Delta$ \\
\midrule
\multirow{4}{*}{Visual-only}
 & Matched embedding & 32.92 & -- \\
 & Text-only SFT & 21.96 & $-$10.96 \\
 & Zero embedding & 20.25 & $-$12.67 \\
 & Shuffled embedding & 21.13 & $-$11.79 \\
\midrule
\multirow{4}{*}{VL-joint}
 & Matched image and question & 65.56 & -- \\
 & Shuffled image & 22.63 & $-$42.93 \\
 & Blank image & 22.01 & $-$43.55 \\
 & Shuffled encoder question & 25.60 & $-$39.96 \\
\bottomrule
\end{tabular*}
\captionof{table}{Controls on the GEB test set. Deltas are relative to the matched condition within each encoding mode. The decoder receives the correct textual question in all VL-joint rows.}
\label{tab:controls}
\end{minipage}
\end{center}

Matched visual embeddings exceed zero and shuffled inputs by 12.67 and 11.79 points. Under VL-joint, perturbing the image or encoder-side question costs 39.96--43.55 points, confirming dependence on matched, sample-specific information.

\subsection{Comparison with MMEB-V2}
\label{sec:mmeb_comparison}

\begin{center}
\begin{minipage}{\columnwidth}
\centering
\small
\setlength{\tabcolsep}{3pt}
\begin{tabular*}{\columnwidth}{@{\extracolsep{\fill}}lrrr@{}}
\toprule
Model & MMEB-V2 & GEB vis. & GEB VL \\
\midrule
Embed-RL-2B & 66.78 & 29.21 & 45.91 \\
VLM2Vec-V2 & 57.49 & 29.26 & 46.09 \\
UME-R1-2B & 59.69 & 31.21 & 51.45 \\
Qwen-Emb-2B & 73.25 & 32.92 & 44.87 \\
Qwen-Emb-8B & 77.82 & 33.21 & 65.56 \\
\bottomrule
\end{tabular*}
\captionof{table}{MMEB-V2 Overall and GEB test scores. Scales differ; Qwen-Emb denotes Qwen3-VL-Embedding.}
\label{tab:mmeb_compare}
\end{minipage}
\end{center}
Table~\ref{tab:mmeb_compare} compares both GEB modes with MMEB-V2 Overall for the five models with archived scores~\cite{meng2025vlm2vecv2}.
Neither GEB mode reproduces the MMEB-V2 ordering. Under visual-only, UME-R1-2B and VLM2Vec-V2 rank above Embed-RL-2B despite lower MMEB-V2 scores. Under VL-joint, Qwen3-VL-Embedding-2B falls from second on MMEB-V2 to last, while UME-R1-2B rises to second. These changes show that GEB provides distinct views beyond the MMEB-V2 ranking.

\section{Discussion}

\textbf{Reusable retention and query-conditioned compression.}
The two encoding modes correspond to two deployment regimes. Visual-only embeddings are reusable: an image is encoded once and served to many future queries that are unknown at encoding time. VL-joint gives up this reusability, re-encoding the image for each question so that representational capacity can be concentrated on the query at hand. This produces substantially larger gains and reorders models relative to visual-only. Reusability and query conditioning, however, need not be mutually exclusive. A lightweight query-conditioned module could adapt a precomputed image embedding to an incoming question, recovering part of the selectivity of joint encoding while retaining offline precomputation and low serving cost. The large gains under joint encoding further suggest that strong vision-language backbones are promising starting points for such modules.

\textbf{Information bottlenecks.}
A fixed-capacity embedding imposes an information bottleneck, and the category results show where recoverability degrades most sharply. Natural-image content is more readily decoded from a compact vector, whereas scene text and visual documents---which hinge on exact strings, numerical values, and layout---are substantially less recoverable. A single embedding budget therefore serves these inputs unequally, motivating representations whose capacity tracks the information demands of each input. Adaptive length, learned embedding slots, and hierarchical summaries offer concrete starting points. GEB can evaluate such designs at both the category and question level, separating broad scene retention from exact-value, text, and layout recovery instead of collapsing them into one aggregate score.

\section{Conclusion}

We introduced the GEB, which evaluates frozen multimodal embeddings through a common generative readout over development and test sets spanning natural images, scene text, and visual documents. Across seven models, visual-only scores cluster tightly, whereas VL-joint changes performance and rankings. Scene text and visual documents remain difficult; controls verify reliance on matched embeddings, and MMEB-V2 yields distinct rankings. Strong retrieval performance therefore does not necessarily imply preservation of fine-grained information needed by downstream generators. GEB complements separability-based evaluation and motivates adaptive-capacity, query-conditioned embeddings.

\noindent \textbf{Generative AI Use Disclosure.}
Generative AI tools were used solely for language editing and presentation. All technical content was verified by the authors.

\bibliography{aaai2027}

\clearpage
\twocolumn[
\begin{center}
    {\LARGE\bfseries Supplementary Material}
\end{center}
\vspace{1em}
]
\setcounter{section}{0}
\setcounter{subsection}{0}
\setcounter{table}{0}
\setcounter{figure}{0}
\setcounter{equation}{0}
\renewcommand{\thesection}{S\arabic{section}}
\renewcommand{\thesubsection}{S\arabic{section}.\arabic{subsection}}
\renewcommand{\thetable}{S\arabic{table}}
\renewcommand{\thefigure}{S\arabic{figure}}
\renewcommand{\theequation}{S\arabic{equation}}
\input{supplementary-content.tex}

\end{document}

%% file: supplementary-content.tex
%
The main conclusions depend on GEB measuring recoverable embedding content rather than artifacts of input construction, an undertrained readout, ambiguous test items, or direct data leakage. This supplement addresses those alternatives through protocol and readout checks, benchmark-construction audits, and uncertainty analyses, then uses a query-mismatch diagnostic to clarify what VL-joint embeddings encode. Protocol choices were made without held-out test results; studies completed before the final GEB split are explicitly separated from final GEB development and test analyses.

\section{VL-Joint Input Construction}
\label{sec:supp_protocol_construction}

VL-joint training requires deciding what text the embedding should represent and what text should remain visible to the decoder. Before the final GEB split, we compared five constructions with Qwen3-VL-Embedding-2B, a Qwen3-0.6B decoder, and full fine-tuning. This preliminary protocol study uses a pre-split QA aggregate and is not directly comparable to the final GEB development scores in the main paper.

Encoding the current question while retaining prior turns as decoder text gives the highest score (79.94). It exceeds full-history encoding with decoder history by 1.20 points and current-question encoding without prior decoder history by 1.65 points. These comparisons favor aligning each embedding with one supervised answer while preserving nonvisual dialogue context. Because the study is not fully factorial, it does not isolate the contribution of each component; we use it only to select the common VL-joint construction.

\begin{center}
\begin{minipage}{\columnwidth}
\centering
\small
\setlength{\tabcolsep}{2pt}
\begin{tabular*}{\columnwidth}{@{\extracolsep{\fill}}llr@{}}
\toprule
Encoder text & Decoder context & QA \\
\midrule
Conversation qs. & History & 75.45 \\
Full history & Embedding only & 76.81 \\
Full history & History & 78.74 \\
Current q. & History & \textbf{79.94} \\
Current q. & Current q. & 78.29 \\
\bottomrule
\end{tabular*}
\captionof{table}{Preliminary VL-joint protocol study. History denotes prior textual turns; Embedding only provides no decoder text.}
\label{tab:protocol_construction}
\end{minipage}
\end{center}

\section{Readout Adequacy Checks}
\label{sec:supp_readout_ablations}

The readout must be capable enough to expose recoverable embedding content while remaining fixed across backends. The main paper's capacity ablation shows that moving from the selected 0.6B one-stage decoder to either 1.7B variant adds only 1.95--2.22 points. We therefore retain the smaller common decoder instead of optimizing decoder capacity per embedding backend. The two checks below address more direct readout bottlenecks: restricted fine-tuning and insufficient training diversity. Both use an earlier protocol-development QA aggregate and are not directly comparable to final GEB development scores.

\paragraph{LoRA and full fine-tuning.}
With the Qwen3-0.6B decoder, Qwen3-VL-Embedding-2B visual-only representation, LLaVA-NeXT training data, and learning rate $10^{-4}$ held fixed, full fine-tuning scores 57.01 versus 37.96 for LoRA. This 19.05-point gap indicates that LoRA would impose a substantial readout bottleneck, so all main comparisons train the adapter and decoder LM end to end.

\paragraph{Training-data scale.}
Holding the optimizer budget at 1{,}440 steps, we then vary the number of unique LLaVA-NeXT conversations. The QA aggregate rises throughout Table~\ref{tab:data_scale}; smaller subsets would therefore make the readout more data-limited under the same training budget. The study does not establish that the full-data setting has reached a plateau, but it motivates using the complete corpus uniformly across embedding backends.

\begin{center}
\begin{minipage}{\columnwidth}
\centering
\small
\begin{tabular*}{\columnwidth}{@{\extracolsep{\fill}}lrrrrr@{}}
\toprule
Unique subset & 25K & 50K & 100K & 200K & Full \\
\midrule
Development QA & 30.68 & 39.90 & 46.66 & 48.17 & \textbf{57.01} \\
\bottomrule
\end{tabular*}
\captionof{table}{Development-stage data-scale study with 1{,}440 steps. Full uses all LLaVA-NeXT 738K conversations.}
\label{tab:data_scale}
\end{minipage}
\end{center}

\section{Benchmark Validity and Release}
\label{sec:supp_construction_details}

To attribute an incorrect answer to inaccessible embedding content, the source item must be in scope, answerable from the image, and unseen by the trained readout. The construction pipeline addresses these requirements before sampling the final development and test sets.

\paragraph{Semantic scope.}
The natural-image category uses all RealWorldQA items; the 2D Count, 2D Relation, and 3D Depth subsets of CV-Bench; and the artwork, landmark, scene, celebrity, commonsense-reasoning, count, color, position, and existence categories of MME after excluding chart-like questions. Scene text uses OCR, posters, and text translation from MME; text-dependent questions from TextVQA; and the regular, irregular, artistic, handwriting, and scene-text VQA subsets of OCRBench. Visual documents use ChartQA, DocVQA, InfographicVQA, and the key-information-extraction subset of OCRBench. These rules are fixed before answerability filtering.

\paragraph{Reference-VLM answerability.}
Category membership alone does not make an item a fair content-recoverability test: some source questions are ambiguous, unanswerable from the image, or depend on external knowledge. We therefore retain only items that Gemini-3-Flash answers correctly under the source task's metric before any GEB-specific filtering. This reduces the chance that a low GEB score is caused by an obviously unanswerable or ambiguously scored item, but it does not establish answerability for every embedding--decoder pair. Because the filter is model-dependent, GEB is an answerable-pool benchmark rather than a model-neutral sample of each source task; Gemini-3-Flash is therefore not reported as a GEB comparator.

The answerability cache is produced with Gemini-3-Flash. Images are converted to RGB, resized to at most 1{,}344 pixels on the longest side, and encoded as JPEG at quality 88. Each question is preceded by: ``Answer the visual question using only the final answer. Use a single word or short phrase. Do not explain.'' Generation uses temperature 0 and a 64-token limit. Predictions are scored with the same LMMs-Eval implementations used by GEB. Binary tasks require a correct score, while TextVQA, DocVQA, and InfographicVQA require at least 0.5 under their consensus or ANLS metric. The cache contains 20{,}237 unique source-task/item pairs, all with successful responses.

\paragraph{Sampling and leakage checks.}
A benchmark built from existing VQA sources risks near-duplicate images crossing the development/test boundary. We assign complete image-identity groups to splits using stratified sampling with seed 20260722. The final audit finds zero overlap in row keys, image-group identifiers, decoded-image hashes, and cross-split perceptual hashes.

The decoder could also memorize test content seen during its own training. Comparing all 900 test items against the LLaVA-NeXT training manifest yields no normalized-path matches; nine candidates identified from filenames or source identifiers also produce no MD5 matches. These checks effectively rule out direct test-data leakage into decoder training.

\paragraph{Release and licensing.}
The planned GEB release uses separate development and test manifests containing source identifiers, category labels, question and answer fields, image-group identifiers, and audit checksums. Where source licenses prohibit media redistribution, reconstruction metadata replaces repackaged files. Upon publication, code will be released under Apache 2.0 and GEB annotations and metadata under CC BY 4.0; source content remains governed by its original license.

\section{Reproducibility and Uncertainty}

The following details fix the shared readout configuration and delimit the statistical evidence behind the main comparisons. Unlike the preliminary studies above, all scores in the uncertainty tables come from the final 900-item test set.

\subsection{Final Decoder Configuration}

Table~\ref{tab:hyperparameters} lists the final readout configuration shared by all main comparisons. Model-specific changes are limited to the input dimensionality handled by the adapter and the two-slot CLIP/SigLIP reference interface described in the Method section of the main paper.

The visual-only training set contains approximately 736{,}900 image--conversation records. In VL-joint mode, expanding conversations by supervised assistant turn produces approximately 4{,}601{,}635 training instances. For models with native image--text joint encoding, each instance contains one fused embedding of the image and current user question. CLIP and SigLIP instead use the paired two-slot layout described in the Method section of the main paper.

\begin{center}
\begin{minipage}{\columnwidth}
\centering
\small
\setlength{\tabcolsep}{3.5pt}
\begin{tabular*}{\columnwidth}{@{\extracolsep{\fill}}ll@{}}
\toprule
Component & Value \\
\midrule
Decoder & Qwen3-0.6B \\
Adapter & LN--Linear--GELU--Linear \\
Adapter width & 1{,}024 \\
Trainable & Adapter + decoder LM \\
Data & LLaVA-NeXT 738K \\
Epochs & 1 \\
Optimizer & AdamW \\
LR & $1\times10^{-4}$ \\
Schedule & Cosine, 3\% warmup \\
Max. grad. norm & 1.0 \\
Precision & bfloat16 \\
Per-GPU batch & 4 \\
Grad. accumulation & 16 \\
GPUs & 8 \\
Effective batch & 512 \\
Max. sequence length & 32{,}768 \\
Loss & Assistant tokens \\
Decoding & Greedy, 1 beam \\
\bottomrule
\end{tabular*}
\captionof{table}{Final decoder configuration. Trainable includes all decoder-LM weights; 32{,}768 tokens is a ceiling with dynamic padding.}
\label{tab:hyperparameters}
\end{minipage}
\end{center}

\subsection{Randomness, Runs, and Uncertainty}
\label{sec:supp_uncertainty}

Each main model--mode row comes from one trained decoder rather than an average over independent reruns. The reported intervals therefore need to distinguish test-item sampling variation from unmeasured training instability.

All decoder training runs set both the model seed and data-order seed to 42, which controls model initialization and training-data order. Training-data subsets used in development also use seed 42. Final development/test construction uses seed 20260722. Each main model--mode row is reported from one final training trajectory; controls reuse that checkpoint where specified in the paper. Failed or restarted jobs are not counted as independent runs, and scores are not averaged across independently trained seeds. The uncertainty intervals below therefore quantify test-item sampling variation -- how much the score would move under a different draw of test items from the same pool -- rather than optimization variation across retrained decoders; a decoder retrained with a different seed could still score outside these intervals.

Confidence intervals resample items with replacement. Marginal score intervals use 10{,}000 bootstrap samples with seed 20260722, and paired control gaps use 20{,}000 samples with the same seed. All reported intervals are percentile 95\% intervals. Paired analyses resample per-item score differences, preserving alignment between the matched condition and its control.

Table~\ref{tab:main_uncertainty} gives descriptive marginal intervals for every embedding backend and encoding mode. Table~\ref{tab:control_uncertainty} gives paired intervals for the control gaps central to the main paper's matched-information claim. Both tables quantify test-item sampling variation from the fixed 900-item pool, not optimization variation across retrained decoders.

\begin{center}
\begin{minipage}{\columnwidth}
\centering
\small
\setlength{\tabcolsep}{2pt}
\begin{tabular*}{\columnwidth}{@{\extracolsep{\fill}}lcc@{}}
\toprule
Model & Visual-only & VL-joint \\
\midrule
CLIP ViT-L/14 & 28.25 [25.39, 31.21] & 4.97 [3.64, 6.44] \\
SigLIP SO400M & 30.58 [27.60, 33.49] & 21.99 [19.36, 24.70] \\
Embed-RL-2B & 29.21 [26.40, 32.15] & 45.91 [42.69, 49.04] \\
VLM2Vec-V2 & 29.26 [26.39, 32.33] & 46.09 [42.90, 49.19] \\
UME-R1-2B & 31.21 [28.28, 34.24] & 51.45 [48.26, 54.68] \\
Qwen-Emb-2B & 32.92 [29.93, 35.97] & 44.87 [41.76, 48.00] \\
Qwen-Emb-8B & 33.21 [30.21, 36.30] & 65.56 [62.41, 68.47] \\
\bottomrule
\end{tabular*}
\captionof{table}{Overall test scores with marginal bootstrap 95\% CIs. Qwen-Emb = Qwen3-VL-Embedding; intervals reflect test-item, not training-run, variation.}
\label{tab:main_uncertainty}
\end{minipage}
\end{center}

\begin{center}
\begin{minipage}{\columnwidth}
\centering
\small
\setlength{\tabcolsep}{1pt}
\begin{tabular*}{\columnwidth}{@{\extracolsep{\fill}}lcc@{}}
\toprule
Condition & Score [95\% CI] & Gap [paired CI] \\
\midrule
\multicolumn{3}{l}{\emph{Visual-only}} \\
Matched & 32.92 [29.93, 35.97] & -- \\
Text-only SFT & 21.96 [19.37, 24.57] & -- \\
Zero embedding & 20.25 [17.73, 22.89] & 12.67 [9.90, 15.42] \\
Shuffled embedding & 21.13 [18.55, 23.72] & 11.79 [9.06, 14.50] \\
\midrule
\multicolumn{3}{l}{\emph{VL-joint}} \\
Matched & 65.56 [62.41, 68.47] & -- \\
Shuffled image & 22.63 [19.86, 25.39] & 42.93 [39.58, 46.28] \\
Blank image & 22.01 [19.31, 24.63] & 43.55 [40.17, 46.87] \\
Shuffled enc. question & 25.60 [22.75, 28.43] & 39.96 [36.42, 43.50] \\
\bottomrule
\end{tabular*}
\captionof{table}{Control uncertainty. Scores use marginal 95\% CIs; gaps are matched-minus-control paired CIs. Enc. = encoder; text-only is trained separately.}
\label{tab:control_uncertainty}
\end{minipage}
\end{center}

Generation is greedy and therefore introduces no sampling randomness. Evaluation scripts retain the LMMs-Eval defaults when no explicit evaluation seed is supplied: Python uses seed 0, while NumPy, PyTorch, and few-shot sampling use seed 1234. The shuffled-embedding control is deterministic: within each evaluation batch, embeddings are displaced by a fixed cyclic shift of one position rather than by random permutation.

\subsection{Software and Hardware}

Experiments were conducted on eight NVIDIA A800 GPUs with 80\,GB memory each under Ubuntu 20.04.6 LTS, using CUDA 12.4, Python 3.10.20, PyTorch 2.6.0, Transformers 4.57.6, Accelerate 1.10.0, PEFT 0.19.1, and LMMs-Eval 0.7.1. DeepSpeed was not installed or used in the training pipeline.

\section{Query-Conditioning Diagnostic}

The main paper's shuffled-encoder-question control shows at test-set scale that VL-joint performance depends on matching the encoder and decoder questions. Table~\ref{tab:wrong_question} clarifies this dependence with four Qwen3-VL-Embedding-8B probes across the two images in the main qualitative figure. Each probe uses batch size one, greedy decoding, and the same VL-joint checkpoint; the decoder receives the target question while the encoder receives a different valid question about the same image.

\begin{center}
\begin{minipage}{\columnwidth}
\centering
\small
\setlength{\tabcolsep}{1pt}
\begin{tabular*}{\columnwidth}{@{\extracolsep{\fill}}llcc@{}}
\toprule
Target & Encoder question & Ref. & Output \\
\midrule
\multicolumn{4}{l}{\emph{Basketball photograph}} \\
Center color & Left jersey no. & White & 14 \\
Behind center & White jersey no. & Hoop/backboard & 23 \\
\midrule
\multicolumn{4}{l}{\emph{Social-media chart}} \\
Twitter \% & Tumblr \% & 44 & 10 \\
TikTok vs. Reddit & Top platform & TikTok & YouTube \\
\bottomrule
\end{tabular*}
\captionof{table}{Selected query-mismatch probes. Top platform asks which platform has the highest share.}
\label{tab:wrong_question}
\end{minipage}
\end{center}

In the basketball probes, asking the encoder for a jersey number produces 14 or 23 even though the decoder asks for a color or a structure. In the chart probes, substituting the Tumblr-percentage or top-platform question similarly produces 10 or YouTube instead of the target answer. In all four cases, the output follows the encoder-side question despite the correct decoder-side text, illustrating why VL-joint deployment requires matched questions. These probes diagnose a mechanism; they are not an estimate of benchmark accuracy.